# BECOS: Consolidating Session Evidence for Coding Teams

**Nikhil Kumar Jha**
becosAI
nikhil@becos.ai

**Amitoj Cheema**
becosAI
amitoj@becos.ai

## Abstract

Developers inherit each other's code, but not the reasons behind it, which stay in past coding-agent sessions. BECOS (Bidirectional Episodic Consolidated Over Sessions) turns those sessions into claims, such as decisions, requirements and findings, each kept with who stated it, whether it was proposed, adopted or tested, and its source session, and gives a later task a small packet of that evidence.

We evaluated BECOS on five benchmarks:

- **Team questions.** On a real team's questions whose answers existed only in past sessions, an agent with BECOS was correct in 8 of 9 runs, against none without memory. Where the answer took a long investigation, it used 82% fewer tokens at 70% lower cost.
- **Mobility.** Over ten sequential coding tasks, the final code passed about 18 of 21 hidden tests with BECOS, against 14 without.
- **RAG in a Trenchcoat.** BECOS scored highest of the memory systems compared on all four stated-rule tests, following a rule buried in 24 sessions in 67% of answers against 30%.
- **LoCoMo.** With a tenth of the text or less, BECOS beat the whole conversation by 7 to 11 points with one answer model and matched it with another.
- **BEAM.** With about 2% of the tokens, BECOS led on all ten abilities under the same prompt, and in 13 of 20 conversations under BEAM's own setup.

Keeping each claim's source and status lets a small packet stand in for a long history, most of all where no note, commit or code holds the answer.

## 1. Introduction

A team builds a mobility service over several coding sessions. In one session, a developer works out which database the service should open. In another, someone adds database migrations. Later, a third developer must make sure that data survives a restart.

The third developer has the code. They may never have seen the earlier discussion. The code shows what the service does today. It does not always show what the team meant it to do. The shared code stays, but the conversation that explains it is gone.

Searching old conversations helps only in part. A search can find a passage about the database that does not say which database the team agreed to use. A summary is cheaper to read, but whoever writes it cannot know what will be asked later: what applies today, why an option was rejected, or whether a change was ever tested.

BECOS carries project knowledge across coding sessions while preserving who stated it, whether it was adopted, what software it concerns, and what evidence supports it. It works in two steps:

- **Remember.** BECOS captures each exchange as the session runs, so nobody has to stop and write a note. At that point the full conversation is available but the future question is not known. BECOS saves claims from it, together with their source and status.

- **Recall.** When a question arrives, only a small part of the history fits beside it. BECOS selects the claims that help answer it.

This paper makes four contributions:

1. A memory for coding teams that keeps each claim with its status in the project, the evidence for it, the software it concerns and who stated it (§3–4).
2. Remember and recall steps that carry all four from the session where a claim was made to the agent that needs it, with no note-writing step (§5–6).
3. A worked example with a real recalled packet from benchmark runs (§7).
4. An evaluation on five benchmarks, including finished code over ten sessions and questions from a real team's work (§8–9).

## 2. Related work

This section places BECOS among memory systems for agents and for coding agents. Appendix A describes each one.

**Memory for agents.** Earlier systems retrieve outside evidence [2], store experiences [3], manage a limited context [4], update facts [5] and track when they were valid [6]. H-MEM, TiMem and MemForest organize memory in layers or trees [8–10]. BECOS adds each decision's status and evidence. Other work shares memory between users [11, 12].

**Memory for coding agents.** projectmem logs decisions and failed fixes [15]. Stratum keeps what agents learned about a repository [16]. CoReMem carries code locations and commands to the next task [17]. VibeMemBench tests whether memory of an agent's earlier fixes helps it solve new issues [18].

**What is different.** Our contribution is the combination, for coding teams: an agent reading a claim can see its status, its evidence, the software concerned and the original session.

## 3. The model in brief

The name describes the design:

- **Bidirectional:** evidence moves from sessions into memory, and from memory into later sessions.
- **Episodic:** each record starts from an episode, which is one exchange at a known time.
- **Consolidated over sessions:** several sessions can support the same claim, and each remains a separate source. A decision can outlive the session in which it was made.

Psychology separates memory of events from general knowledge [1]. A team needs both. “On Tuesday the team agreed to X” is an event. “X is the rule” is knowledge. BECOS keeps the event as the evidence for the rule. We borrow the distinction, not a theory of how human memory works.

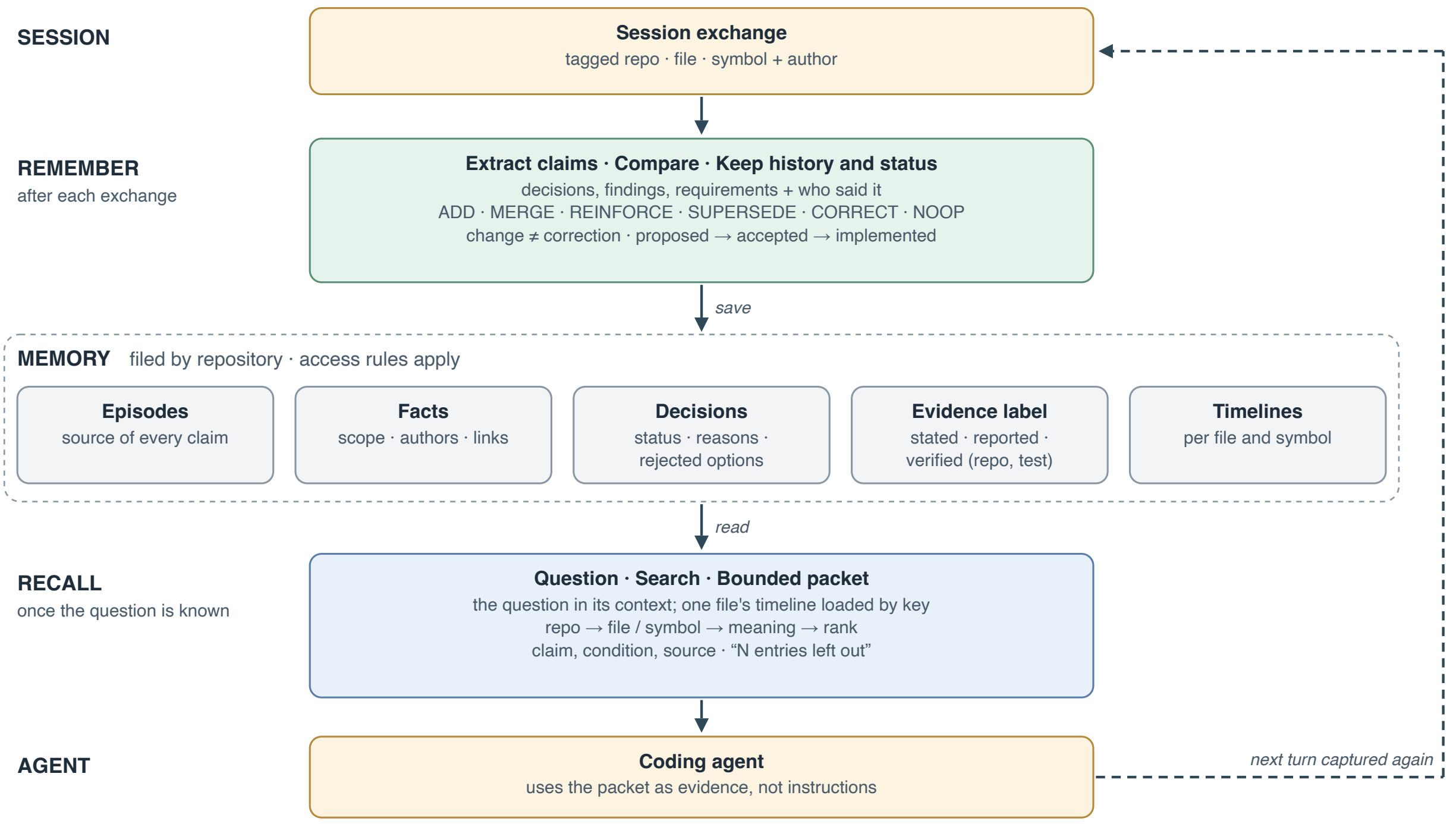

Figure 1. A session exchange (top) is tagged with the repository, file, symbol and author it concerns. Remember (§5) extracts claims with who said them, compares each with a few related claims and gives it one of six update actions, and keeps its history and status. Memory (§4) files every claim by the repository it is about: episodes, facts, decisions with their status, evidence labels kept apart from status, and a timeline per file and symbol. Recall (§6) reads the question in its context, searches and builds a bounded packet that says what it left out. The coding agent (bottom) uses the packet as evidence, and its next turn is captured again (dashed).

The design has three goals:

- **Continuity.** A later agent can use what the project learned without rereading the original conversation.
- **Accountability.** A reader can check why memory says something, and whether it was observed or only said.
- **The right use.** Evidence about the past must not quietly become an instruction for today.

Access rules apply throughout. A relevant episode that the reader may not see is never used.

## 4. What memory stores

Memory holds five kinds of object. They can be built with different storage and search technologies.

| Object | What it is | What to keep in mind |
|---|---|---|
| Episode | One exchange, with its session, time and any observations | When something was said is not always when it applies |
| Candidate fact | A claim taken from an episode | It is a first reading and may not be kept |
| Consolidated fact | A reusable claim, with its scope, evidence and sources | Repeating a claim does not verify it |
| Decision | A choice, with its status, reasons and rejected options | Agreed, built and tested are three different things |
| Recall packet | The claims and sources selected for one question | It is evidence for the agent, and the agent still writes the answer |

A *fact* here is a record of a claim, with a source that can be checked. Suppose a developer says that data must survive a restart. That is strong evidence of what they asked for, even if nobody has tested it. If someone then restarts the service and the record is still there, that is evidence of one case. It does not show that every migration is safe. A single confidence score would hide this difference.

A claim has two kinds of boundary. **Access** says who may see it. **Applicability** says what it is about: a project, a file, a group or a condition. Being allowed to read another service's migration rule does not make that rule apply to this one.

**Filed by what it is about.** One session often touches several repositories, so the folder an agent works in is a poor guide to what a claim concerns. The coding agent tags what it says with the repository, file and symbol concerned, and each claim is filed under the repository it names. A claim without a tag in a session about one repository takes that repository. Files and symbols are named within their repository, so the same name in two repositories never refers to one thing.

**Who said it.** Each claim records the person who first stated it, by their sign-in identity, and every person who later stated it again. Recall shows these names beside the claim. One memory can then hold the work of several developers, and a reader can tell whose requirement a rule is.

## 5. Remember: from a conversation to saved claims

Each exchange passes through one write pipeline. It is recorded, claims are extracted from it, related saved claims are retrieved, and each new claim is given one of six update actions: ADD, MERGE, REINFORCE, SUPERSEDE, CORRECT or NOOP. In the deployed system, extraction, comparison and settling use DeepSeek V4 Flash. The steps below describe it.

**Step 1. Record the episode.** BECOS records enough of the exchange to understand it. “Yes, the environment wins” means little alone. It becomes a requirement next to the question it answers: “Which source sets the database path when both are present?” BECOS keeps apart what people said, what was reported as done, and what was observed. Text that came from an earlier recall is not new evidence, and showing it to an agent again does not make it a second source.

**Step 2. Extract claims.** The extractor identifies what could be useful later: a decision, requirement, discovery, change, failed attempt or open question. It writes each claim so that it makes sense on its own, keeping words such as *not* and *only*, who said it, and whether it was *proposed* or *accepted*. It attaches the source and, where useful, a short note on the situation. A reason stays with its decision when the two are clearly linked, and claims stay separate if one could change without the other. Some exchanges add nothing, and then nothing is saved.

**Step 3. Compare before saving.** BECOS looks for earlier claims about the same subject. Similar wording is a reason to look, but not enough to merge. “Apply migrations at startup” and “apply migrations **only** at startup” sound alike, but only the first allows a separate migrate command. Two people can also state the same decision in different words. Only a few related claims are compared, never the whole memory. The new claim is then added as new knowledge (ADD), recorded as another source of a claim already saved (MERGE), counted as reaffirming an existing decision or constraint (REINFORCE), recorded as replacing or narrowing an earlier one (SUPERSEDE), recorded as showing that an earlier claim was wrong (CORRECT), or dropped because it adds nothing lasting (NOOP). When the relationship is unclear, BECOS keeps both, each with its source.

**Step 4. Keep the history.** Consider two statements:

- “I misspoke yesterday; we never adopted automatic migration at startup.”
- “Starting today, the service no longer migrates at startup.”

Both give the same rule today. Only the second means that the old rule once applied. Overwriting the old record would lose that. Keeping both as current would confuse the reader. So a correction and a change are recorded differently.

**Step 5. Settle what happened to a proposal.** An agent often offers options, and the user answers in the next turn. So a decision starts as *proposed* when the agent suggested it, *accepted* when the user chose or asked for it, or *implemented* when the same turn edited files or committed. Later turns of the conversation can move it to *accepted*, *implemented* or *rejected*; it can also be replaced, corrected or reverted. A restatement can move a decision forward, never back. Turns can be processed out of order, so settling works both ways: a reply settles earlier open decisions, and a new proposal looks for a later reply that has already been saved. A model judges each turn, and fixed checks limit what it can decide. *Implemented* needs an edit or a commit in that turn. *Accepted* and *rejected* need the user's own words in that turn.

**Status and evidence are separate labels.** A decision's status says where it stands in the project. Its evidence says why to believe it: stated by a user, reported by an agent, verified in the repository, verified by a test, or a mix. The two are stored separately and never merged into one score. An agent's report that something works stays *reported by an agent*; it is not promoted to verified.

## 6. Recall: from a question to useful evidence

**Step 1. Understand the question.** Recall first settles three things. *Scope:* recall is limited to the repository the session works in or, in a team deployment, covers the team's memory and ranks the session's own repositories higher. *Wording:* when the client sends the last few turns, a small model (Qwen 3.6 27B) rewrites the question so that words such as *that rule* name what they refer to. *Type:* fixed keyword rules sort every question into one of four types, checked in this order:

- **Negative** ("what did we reject or try?").
- **Historical** ("what did we do before?", "what replaced it?").
- **Rationale** ("why?").
- **Current** ("what do we use now?"), the default.

A rewritten question is checked again; this can move it into the historical type but never out of it. Keyword rules are cheap and predictable.

**Step 2. Gather candidates.** Searching by meaning connects "the record is still there after a relaunch" with "persistence across restart": the question is embedded and compared by cosine similarity with the saved claims in scope, and the closest are kept. The reported results use voyage-code-4 embeddings; voyage-code-3, Qodo-Embed-1 and Nomic Embed Code gave equivalent results. A file or symbol that the question names is matched exactly, and every claim about it is added, however it is worded. Decisions are gathered separately: a current or rationale question sees only decisions still in force, a historical or negative one sees all of them, and a decision that was only proposed is never included. Claims the agent already has in its context are dropped, and a claim nearly identical to one already chosen (cosine similarity of 0.90 or more) is folded into it. Access rules apply throughout.

**Step 3. Score and choose.** Each claim's score starts from its cosine similarity and is then adjusted by fixed amounts based on the question type. No learned ranker is used.

Decisions are scored on their own, by cosine similarity plus a quarter of the share of the question's words they contain, and the best five fill a decisions section. For a historical question, the earlier version of each retrieved claim is added and shown as an old → new pair. Together these give each type of question the evidence it needs:

| Question | Most useful evidence |
|---|---|
| What is the rule now? | Decisions still in force, ranked above similar proposals |
| What did we do before? | Replaced decisions, each paired with what replaced it |
| Why? | Decisions with their reasons |
| What did we reject or try? | Rejected approaches and failed attempts |
| Does it work? | No type of its own; claims verified in the repository or by a test rank higher |

Recall does not judge whether a newer remark replaces an older decision; that is recorded when the claim is saved (§5), and at recall recency adds at most 0.05. Some questions leave something open: "Does the migration need to run?" does not say which database the service opens, so the useful answer is a rule with its condition. Because the condition is saved with the rule (§5, Step 2), recall returns both.

**Step 4. Build a small packet.** The packet includes the claim, any condition that changes its meaning, and a link to its source. The design removes repeats before an exception or a disagreement, and if a rule and its exception do not both fit, it says less or marks the point as uncertain. The packet can be checked on its own, so we can see whether the right evidence reached the agent even when its answer was wrong. The packet has a fixed size: 8,950 characters of recalled text, about 9,500 with the header the agent receives. When entries do not fit, it says how many were left out, so the agent knows that more exists and a reader can see what the limit cost (§7).

**The history of one file.** Some questions ask for everything recorded about a file or symbol, not the few claims closest to a question. For these, recall loads that file's or symbol's timeline directly, without searching. Each entry's status comes from the saved claims and decisions, so a later replacement or correction shows up even on an old entry. Near-duplicate statements are shown as one line, but only when they share a status: a *proposed* decision is never folded into an *implemented* one, so a merged line's label is true for every claim it cites. Entries are shown newest first, grouped into constraints and warnings, decisions, other current claims, and replaced or corrected claims, each naming what replaced it.

## 7. Worked example: keeping data across a restart

This example comes from the mobility study runs in which every task starts a fresh session without a handoff note. One backend is built in ten tasks. Each task starts in a new agent session with its prompt, the project's contracts and the code so far. No one writes a handoff note. One setup, without BECOS, has only the code. The other, with BECOS, also saves and recalls session evidence. This is like one teammate handing a feature to another: the conversation changes, but later work must fit the earlier code.

| Handoff | What the next session needs to know |
|---|---|
| Configuration and clock (S03–S04) → storage | The rule for choosing between configuration sources, which is not the same as the value used in one run |
| First migrations (S05) → more storage work (S06) | Which database to migrate, how to handle duplicate or out-of-order versions, and whether one good first startup means that later startups are safe |

**The task.** In the S06 restart task, an earlier session has created a database and applied migrations in order. After a restart, a record saved earlier must still exist exactly once, and a migration already applied must not run again. Starting with an empty database can look fine and hide both problems.

**What good recall would give.** It would say which setting selects the database. It would also separate "startup worked on an empty database" from "a restart kept existing data." The agent could then run the migration twice on the right database.

**What reached the agent.** Each memory run logged the text it was given, so we read the S06 packet of all ten runs (7.4–7.9 thousand characters each).

- Every packet listed facts or decisions with a status, an author and a source identifier, and many entries named the file concerned.
- Every packet said how many further entries it had left out for space (7–12), so the agent knew that more existed.
- Every entry carried its evidence label, stated by the user or reported by the agent, so the agent could tell a requirement from a report.

These entries from the packet of round 1 show database-path resolution, the chosen migration approach and the persistence requirement. The first is shown in full, as the agent received it.

```
Decisions (current):
- [repo:mobility-first10-small path:mobility/config.py] [2026-11-23 10:35:49]
  Relative `database_path` values resolve against the config file's
  directory, or the current working directory when no config file is
  supplied. [implemented] [lifecycle:implemented]
  [scope:branch=master,symbol=resolve_settings] [evidence:agent_reported]
  [occurrences:1] [author:bench-agent@bench.invalid]
  (cite: decision_9515e6330668) (fact: fact_b781c148a4ab4ce3)
  reason: Required to conform to `.pilot-task/adapter-contract.json`
  context: Path resolution rule adopted to match the adapter contract
- [path:mobility/cli.py] `serve` runs migrations before binding the server
  so repeated startup is safe and idempotent. (other options: migrating
  only via the explicit `migrate` command) [implemented]
  [evidence:agent_reported]
Relevant memory (current):
- [user_constraint] The mobility backend must use durable SQLite storage
  with ordered versioned migrations, safe repeated startup and explicit
  migration diagnostics, and must preserve data across service restarts.
  [evidence:user_stated]
[12 recalled entries omitted by output budget]
```

*Excerpt from the S06 packet of round 1. The first entry is complete; in the other two, dates, identifiers and some labels are removed for space. The author is the benchmark agent's sign-in identity.*

Each entry says what was decided, where in the code, its recorded implementation status, who it came from and whether a user stated it or an agent reported it. The two identifiers lead back to the saved decision and its source exchange. The second also names the option that was not chosen. Section 9.1 shows the outcome: the final restart test failed in three runs with memory and eight without.

**What comes back.** A restart that keeps the record is new evidence, for that configuration. The design can save this observation for the next session without turning one good run into a general promise.

## 8. How BECOS was evaluated

We use five benchmarks, because no single one tests every way that past sessions can help. Each asks a different question, uses its own data, and compares BECOS with alternatives on that same data. Scores are never averaged across benchmarks.

| Benchmark | Question | Data | Compared with | Measure |
|---|---|---|---|---|
| Mobility (9.1) | Does memory improve code that continues earlier work? | One synthetic backend, ten tasks | The same agent without memory | Hidden tests passed |
| Team questions (9.2) | Can an agent reuse what a team already worked out? | Ten questions from real sessions | No memory; redoing the work | Correct answers; tokens; cost |
| RAG in a Trenchcoat (9.3) | Does the agent follow what a developer said earlier? | Public, 120 scenario evaluations [13] | Four memory systems | Rule or habit applied |
| LoCoMo (9.4) | Can a long conversation be recalled in little space? | Public, 1,540 questions [7] | Whole conversation | Answers judged correct |
| BEAM (9.5) | Does that hold for longer conversations and more kinds of question? | Public, 20 conversations, 400 questions [19] | Whole conversation; BEAM's search | Rubric score, ten abilities |

In all five, BECOS starts with a fresh or frozen memory that holds only the history the benchmark allows. The BECOS backend used for testing was frozen throughout the benchmarking: no change was deployed to it while a benchmark was running. Each benchmark used the BECOS build of its time, so the results cover several builds and no single release. Section 11 states their limits.

## 9. Results

### 9.1 Mobility: continuing earlier work

**What it asks.** A backend is built in ten dependent tasks, S01–S10. The code carries over from task to task. Does remembering the earlier sessions help the agent finish with more of the system working?

**How it was run.** The agent was OpenCode with DeepSeek V4.1 Flash at low reasoning effort, served by CoreWeave (fp8). After each task, hidden tests were run offline and never shown to the agent. The score is how many of 21 tests the final code passes. The setups being compared ran side by side in each round, so every comparison is paired by round. Each memory run began with an empty memory, and recall was added automatically to each task prompt. Three ways of continuing work were tested:

- **Continuous session:** one conversation for all ten tasks, with automatic context compaction.
- **Fresh session, no handoff:** a new conversation for each task. Only the code carries over.
- **Fresh session with handoff:** each session also receives a note written by the previous one.

The continuous-session and handoff setups were run first, in two sets of rounds. The fresh-session setup without a handoff was run later, in a set of its own. Of its eleven completed rounds, one (round 9) was excluded because of an error at the memory backend's infrastructure provider, which made that round's results invalid; the remaining ten are reported.

**Result.**

| How work continued | Pairs | Without BECOS | With BECOS | Gain [95% CI] | *t* / exact *p* |
|---|---|---|---|---|---|
| Continuous session | 10 | 14.4 | **17.7** | +3.3 [+0.8, +5.8] | .015 / .023 |
| Fresh session, no handoff | 10 | 14.6 | **18.0** | +3.4 [+1.1, +5.7] | .008 / .014 |
| Fresh session with handoff | 9 | 17.8 | **18.2** | +0.4 [−0.7, +1.5] | .38 / .52 |

BECOS added more than three tests in both settings where no note was written. The two results come from separate runs and are close to each other. In the fresh-session runs without a note, BECOS scored higher in nine of ten rounds:

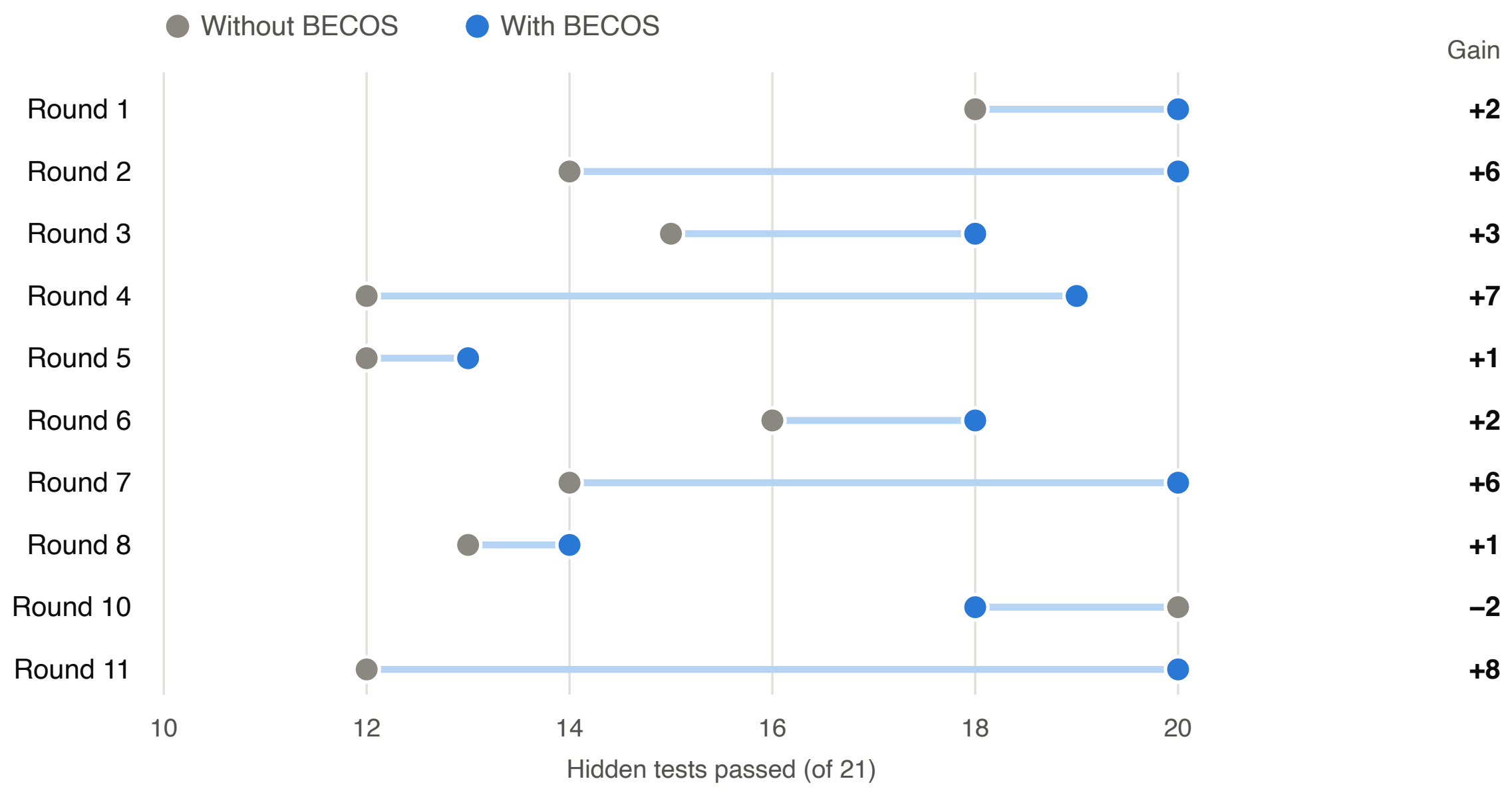

Figure 2. Fresh sessions without a handoff note: hidden tests passed by the final repository in each round (mean 14.6 without BECOS, 18.0 with).

The gain did not come from spending more on the agent. In those runs the agent's billed cost per run was the same ($0.33 in both setups, $p$ = .75; the memory service's own model calls are not included). It made 34 fewer model calls per run on average (−9.5%, $p$ = .010). Agents also got more steps right the first time: 7.1 of 10 steps were fully correct when graded, against 5.8 ($p$ = .039, exact $p$ = .062).

**Where memory helped.** The largest effect was at database migration (S05). Five of the 21 tests depend on the migration command working on a new database. At the end of those runs it was still broken in eight of ten runs without memory and two of ten with it. Restart behavior (S06), the example in section 7, follows from the same step: its final test failed in eight runs without memory and three with it. This fits a later session being told what an earlier one had settled about storage and configuration. One later test moved the other way: inactive-vehicle selection failed in seven memory runs and four baseline runs, for reasons not yet diagnosed.

**Compared with a handoff note.** In the two earlier sets, the continuous-session and handoff setups ran side by side, so memory and a note can be compared round by round. All four ran in nine rounds:

| Comparison (same nine rounds) | Mean difference | 95% CI | Exact $p$ |
|---|---|---|---|
| Note, no BECOS, against a continuous session with neither | +3.3 | [+0.5, +6.1] | .031 |
| BECOS, no note, against a continuous session with neither | +2.9 | [+0.3, +5.5] | .047 |
| BECOS alone against a note alone | −0.4 | [−1.8, +1.0] | .59 |

BECOS alone came within half a test of a note, and the interval rules out a note being better by more than about two tests. The note setups also start a fresh session for every task, so this compares two whole ways of continuing work. With both memory and a note, the score was 18.2. Memory and a note seem to carry much of the same information. In the second of the two sets, memory on top of notes did reduce model calls by about 15% (60 fewer per run, $p$ = .02). A note also needs a closing turn in every session, and in that set the note-writing design cost about 35% more as billed than one continuous session (the same with cold caches). BECOS captures sessions without that turn, and adding it to a continuous session did not raise cost. So memory gives most of a note's benefit without anyone having to write one.

**What it shows.** On this task sequence, memory of earlier sessions improved working code at the same agent cost, and where the two were run in the same rounds, it did about as well as a handoff note written in every session.

### 9.2 Team questions: reusing what a team worked out

**What it asks.** A team has already investigated something. Later, a teammate's agent gets the same question. Does memory give a correct answer, and how much repeated work does it avoid?

**How it was run.** The ten questions come from real work by the authors' team. The memory is a frozen export taken on 31 October 2026, holding 467 captured agent turns (10 August to 31 October) and 13,141 extracted facts, mostly about agent plugins, server code and planning notes. It combines the real work of two developers: 9,569 facts come from one developer's sessions and 3,572 from the other's. Three of the questions go back to investigations by the first developer and seven to investigations by the second. An agent then asked each question again, as a teammate might. For each question, the memory was cut at the moment the question was first asked and the code was set to the same moment, so neither held later answers. Claude Code ran on Sonnet 5.5 with read, search and git tools, in two or three setups, three runs each:

- **With memory:** recall added to the prompt.
- **Without memory:** the code only.
- **Redo:** no memory; the agent first repeats the original investigation, then answers.

Each answer key lists checkable facts, verified in the code or in the session that made the decision and approved by a person before any run. A separate small model rewrote every answer to hide its source, and Claude Opus 5.5 graded it. An answer is correct only when every fact is present. Up to three neutral follow-ups were allowed. Cost and tokens are counted up to the first correct answer, and a saving is measured against the cheapest other setup that also answered correctly.

**Result.**

| Where the answer lives | Questions | With memory | Without memory | Redo | Tokens saved | Cost saved |
|---|---|---|---|---|---|---|
| Only in past sessions | 3 | **8/9** | 0/9 | 0/6 | none to compare | none to compare |
| In a long investigation | 3 | **9/9** | 4/9 | 9/9 | **82%** | **70%** |
| In commits and logs | 2 | 6/6 | 6/6 | not run | 5% | −5% |
| Directly in the code | 2 | 6/6 | 6/6 | 3/3 | −15% | −24% |

**Only in past sessions.** The three questions ask about a cause found by experiment, a plan that was never built, and an approach rejected in discussion. With memory the agent was right in 8 of 9 runs. Without memory it was right in 0 of 9, and in 0 of 6 when it first repeated the investigation. Here memory decides whether the agent has the answer at all. In one of these questions, a developer had worked out in a session how the installer should handle an expired sign-in: a check that asks the server whether the saved sign-in still works, and a new browser sign-in only when it is rejected. The plan was never built, so no code or document held it. When the question came up again, as a teammate would ask it, the agent with BECOS gave the plan in all three runs, and the agent without memory in none.

**In a long investigation.** The answer can be rebuilt from the code, at a price.

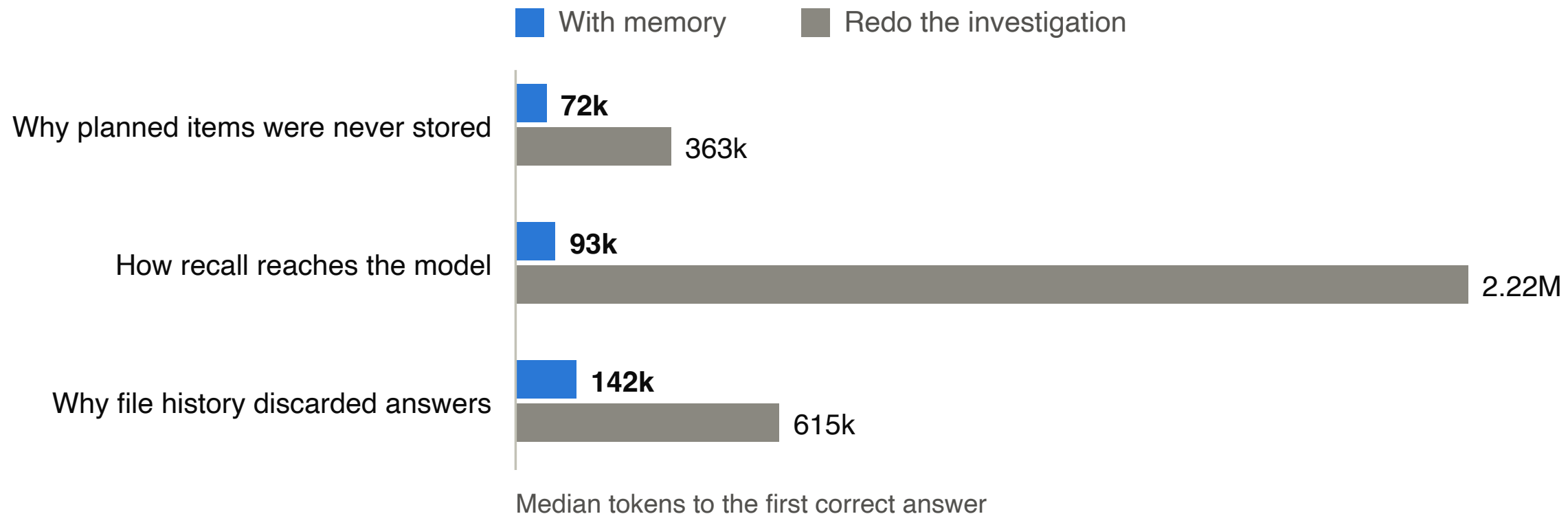

Figure 3. Median tokens to the first correct answer. Median cost was $0.10, $0.13 and $0.14 with memory, against $0.26, $1.08 and $0.51 for repeating the investigation.

Memory was correct in every run. Repeating the work was also always correct, but it cost 2.6 to 8.1 times as much and used 4 to 24 times the tokens. Without memory or a full repeat, agents reached a correct answer in four of nine runs. On the third question they never did: they kept the faulty rule or proposed another that would not work. The savings in the category table use the cheapest correct alternative, which for the first two questions was the setup without memory.

**Already written down.** When the answer was in a commit, a log or the code itself, every setup found it in every run, so there was little left to gain. With memory, agents used 5% fewer tokens when the answer was in commits and logs, and 15% more when it was in the code.

**What it shows.** Memory is most valuable where other sources are weakest: for decisions and findings that were discussed but never written down, and for investigations that cost a lot to repeat. The questions and per-run results are public at becos.ai/benchmark; the sessions behind them are private.

### 9.3 RAG in a Trenchcoat: following what a developer said

**What it asks.** A developer states a rule once, such as never storing money as floats. Sessions later, nobody repeats it. Does the agent still follow it? This public benchmark [13] hides a rule or a demonstrated habit in past conversations, then gives a request that does not mention it. It is also our only direct comparison with other memory systems on one set of scenarios.

**How it was run.** We ran all five original probes and the harder rule condition: six conditions, 20 synthetic scenarios each. BECOS captured each scenario's sessions in its normal way, in date order, with a separate memory per scenario. Answers used gpt-4.1-mini, the answer model of the original paper, three times per condition and six for the dialogue. The other systems' scores come from the original authors' saved answers. Every answer, ours and theirs, was graded by the same Gemini 3.1 Pro grader, which sees the request, the rule and the answer but never the memory. *Bridge* keeps conversation blocks, including raw text; *perfect information* gives the model the source of the rule and marks the ceiling.

**Result.**

| Test | BECOS | LangMem | Graphiti | Mem0 | Bridge | Perfect information |
|---|---|---|---|---|---|---|
| Follow a rule stated once | **83.3** | 80.0 | 70.0 | 40.0 | 35.0 | 95.0 |
| Respect a ban stated once | **91.7** | 90.0 | 65.0 | 75.0 | 60.0 | 95.0 |
| Follow a rule buried in 24 sessions | **66.7** | 30.0 | – | – | 30.0 | – |
| Keep a rule through a dialogue | **78.3** | 75.0 | 60.0 | 55.0 | 35.0 | 95.0 |
| Repeat a procedure only shown | 62.6 | 56.6 | 51.5 | 53.5 | **85.9** | 90.9 |
| Follow a code style only shown | 43.9 | 46.5 | 75.0 | 45.1 | **79.7** | 94.7 |

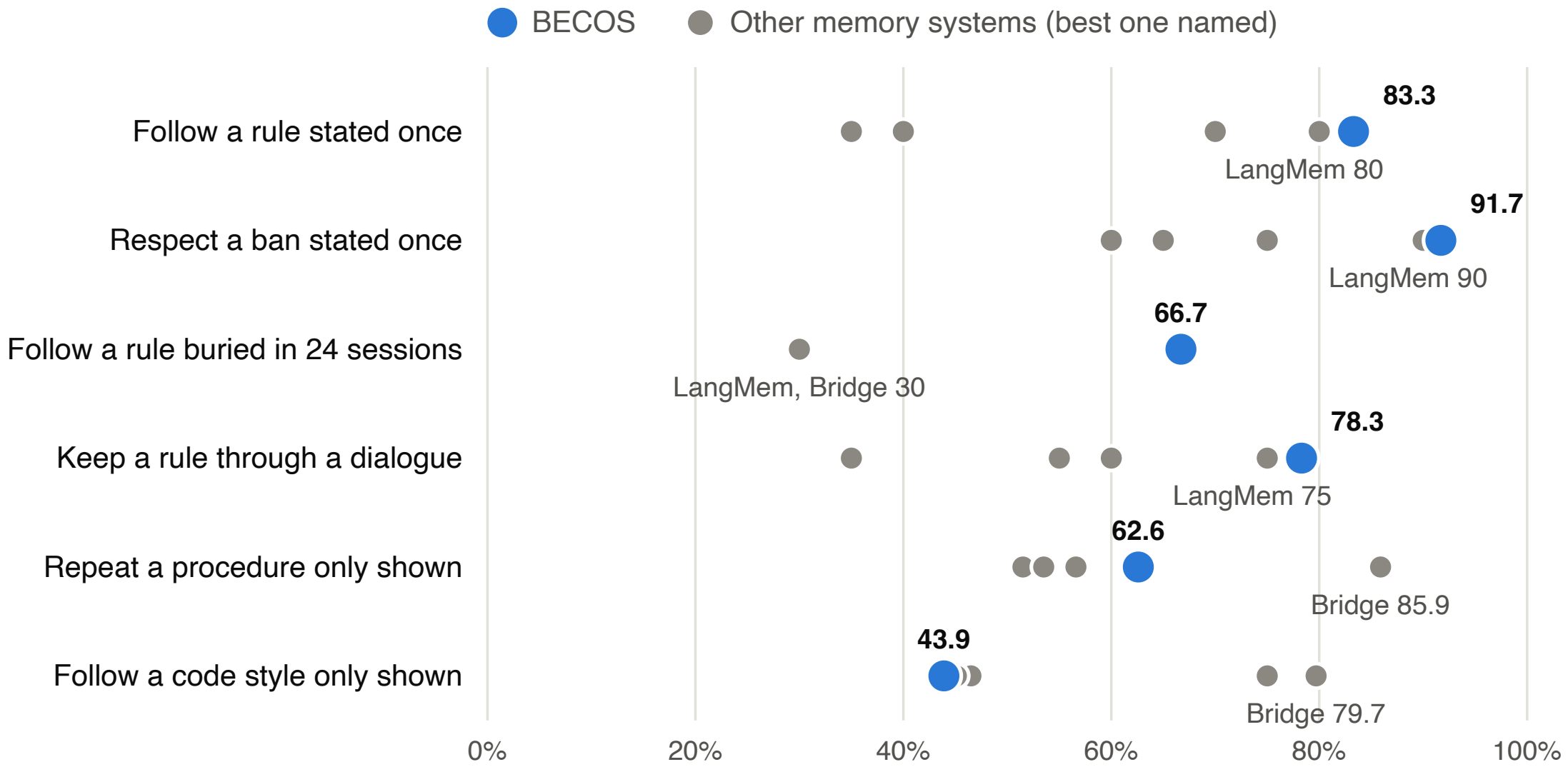

Figure 4. The same scores. The harder rule condition has saved answers for two other systems only.

**Rules.** BECOS had the highest score of the memory systems with saved answers in all four conditions about something a developer said. The clearest result is the hardest one. When a rule was stated in session 3 of 24 and later sessions covered similar topics, BECOS followed it in 66.7% of answers, against 30% for both systems with saved answers. On the simpler tests BECOS and LangMem are close.

Measuring recall separately explains the rule scores. In the first test, the rule was in the text given to the model in 19 of 20 scenarios; in the buried condition, 18 of 20. The remaining misses are therefore mostly the answering model not applying a rule it had been shown.

**A rule through a dialogue.** Here a simulated developer follows the first request with four more. BECOS kept the rule in 78.3% of dialogues. At every turn the rule was retrieved in 52 to 57 of the 60 dialogues checked. Searching with only the newest message, as a coding agent's plugin does, retrieved it at least as often as searching with the whole conversation.

**Habits that are only shown.** The last two tests contain no statement to remember. The agent must infer a procedure or code style from examples. BECOS was second on procedures, ahead of LangMem, Graphiti and Mem0. On code style it was level with LangMem and Mem0, and Bridge and Graphiti, which retain more of the original material, did better. This fits the design. BECOS records decisions, constraints and their reasons. A style visible only in earlier code is a pattern that no one stated. A coding agent can normally read style from nearby files, which this test withholds.

**What it shows.** Rules that a developer states can be saved with a source and status and recalled later, even when buried among many similar sessions.

### 9.4 LoCoMo: long conversations in little space

**What it asks.** LoCoMo [7] tests memory of long personal conversations. It is not about coding teams; we include it because it is widely reported and tests how well a small context replaces a long one. Full results are in a separate paper [14].

**How it was run.** Ten conversations and 1,540 questions, with two answer models, two answer prompts and three judges. The whole conversation was rerun as a baseline under identical conditions. BECOS used a research configuration for conversations, frozen before the runs, and gave the model 8,774 characters on average against 89,048 for the whole conversation: 90% fewer. The score is the share of answers judged correct by gpt-4o-mini.

| Answer model | Answer prompt | No memory | Whole conversation | BECOS | Difference [95% CI] |
|---|---|---|---|---|---|
| gpt-4o-mini | Short | 19.5 | 73.0 | **79.7** | +6.8 [+3.8, +9.5] |
| gpt-4o-mini | Seven-step | – | 77.5 | **88.8** | +11.2 [+9.2, +13.1] |
| gpt-4.1-mini | Short | – | 83.1 | 83.2 | +0.1 [−2.3, +2.2] |
| gpt-4.1-mini | Seven-step | – | 92.1 | 92.5 | +0.3 [−0.6, +1.1] |

**What it shows.** With gpt-4o-mini, BECOS is clearly better than giving the model everything. With gpt-4.1-mini it matches the whole conversation while using a tenth of the text. The gain with gpt-4o-mini is largest on questions about time (79.8 against 50.5 with the short prompt). A blind human check of 150 answers found that this judge's errors on disputed answers worked against BECOS.

### 9.5 BEAM: longer conversations, more kinds of question

**What it asks.** BEAM [19] is a newer public benchmark of long conversations between a user and an assistant. Each conversation has 20 questions covering ten abilities: abstention, contradiction resolution, event ordering, information extraction, instruction following, knowledge update, multi-session reasoning, preference following, summarization and temporal reasoning. A judge model scores each answer against a rubric, from 0 to 1. It asks whether a small context can replace a long conversation across more kinds of question.

**How it was run.** We used the first 20 conversations of the 100K-token split, with BEAM's own code for prompts and scoring and the chat files that code loads. Answers used gpt-4o-mini, and gpt-4.1-mini judged them, the judge in BEAM's code. Each question was answered four times. BECOS stored each conversation in order into a fresh memory and gave the model about 2,200 tokens per question. The whole conversation was 94,000 to 123,000 tokens, with nothing cut.

BEAM's code gives the two kinds of setup different instructions. The whole conversation gets a short-answer prompt. Memory and search get a stricter one: answer only from the given context, without the model's own knowledge. We report BECOS both ways: with BEAM's setup, and with the same prompt as the whole conversation, which changes the instructions and nothing else. Conversations 10 to 20 were run both ways, with settings fixed before the runs. Conversations 1 to 9 were run earlier, with BEAM's setup only.

| Setup (conversations 10–20, 220 questions) | Prompt | Context (tokens) | Score | Difference | Conversations ahead |
|---|---|---|---|---|---|
| No memory | Short-answer | – | 0.133 | | |
| Whole conversation | Short-answer | 116,900 | 0.376 | | |
| BECOS | Short-answer (same) | 2,100 | **0.485** | **+0.109** | 8 of 11, level in 2 |
| BECOS | BEAM's strict prompt | 2,200 | 0.406 | +0.030 | 6 of 11 |

With the same prompt, BECOS scored higher than the whole conversation in all four answers on 56 of the 220 questions, and lower in all four on 18. It was ahead on all ten abilities. The largest gains were on preferences (0.79 against 0.53), information extraction (0.67 against 0.51), knowledge updates (0.44 against 0.30) and temporal reasoning (0.18 against 0.09). On conversations 1 to 9, with BEAM's setup, BECOS scored 0.463 against 0.406 and was ahead in 7. Over all 20 conversations with BEAM's setup, it was ahead in 13.

We also ran BEAM's own retrieval method on conversations 1 to 8: each user–assistant pair is a document, and the five closest pairs are retrieved [19]. All three setups below used BEAM's strict prompt.

| Setup (conversations 1–8, 160 questions) | Context (characters) | Score |
|---|---|---|
| Search, top five pairs (BEAM's method) | 23,700 | 0.410 |
| Search, same space as BECOS | 8,950 | 0.378 |
| BECOS | 8,700 | **0.459** |

**What it shows.** With the same instructions, a context of about 2% of the tokens beat the whole conversation, in 8 of 11 conversations and on every ability. With BEAM's own setup BECOS was also ahead: by 0.030 on conversations 10 to 20, and in 13 of all 20 conversations. The strict prompt alone cost BECOS 0.079 on the same recalled text. Against BEAM's own retrieval method, BECOS scored higher with about a third of the text. Reading the questions BECOS missed points to two improvements. In two questions the recalled text held both the user's update and a later assistant restatement of the old value, and the answer model sometimes chose the old one; marking which value replaced which would settle these. In four conversations, a user's standing formatting rule was not among the recalled claims.

## 10. Discussion

**Memory helps most where nothing else holds the answer.** The benchmarks differ in data, models and measures, but they agree on this. Questions whose answers existed only in past sessions were answered with BECOS and by no other setup. In the mobility study, both settings without a written note gained more than three tests, and memory alone did about as well as a note. Where a commit, the code or a handoff note already held the answer, every setup found it, so there was little left to gain.

**Stated rules carry over.** BECOS saves claims with their status and evidence (§4), and the Trenchcoat results fit that design. Rules and bans stated once were recalled later, even among many similar sessions.

**A small context can be enough.** On LoCoMo, a tenth of the text matched the whole conversation with one answer model and beat it with another. On BEAM, about 2% of the tokens scored higher than the whole conversation on all ten abilities when both had the same instructions. On long investigations, agents with memory reached correct answers with a fraction of the tokens needed to repeat the work. Both fit the idea of organizing evidence once, so that each later reader does not have to.

**One team, one memory.** The team questions use one memory built from two developers' real work, and the expired sign-in plan in §9.2 shows the point: work done in one developer's session reached a later agent although it never became code. In the mobility tasks, new agent sessions stand in for teammates. What remains to be tested is several people writing code together through one memory. BECOS is built for that setting, and it is the next experiment: one developer states a requirement, another observes a result, and a third makes a change that must respect both.

## 11. Limitations and responsible use

**Limits of each benchmark.**

- **Mobility.** New sessions stand in for teammates; no human teams were observed, and each run had one automated author. There is one task sequence, one model and one provider, and effects may differ elsewhere. Several hidden tests depend on the same migration behavior. One round of the fresh-session runs without a note was excluded for an infrastructure error (§9.1). Confidence intervals and $p$-values are nominal and not adjusted for multiple comparisons. No other memory system, and no plain search over the raw sessions, was run on these tasks. Recall settings differed from product defaults.
- **Team questions.** There are ten questions with three runs each. They come from the authors' own sessions, which are private. The work was real, but an agent asked each question again rather than a teammate.
- **Trenchcoat.** Each test has 20 synthetic scenarios, with intervals of roughly 15–20 points. The other systems were not rerun. The recall change in the BECOS build used was found while studying the rule tests, so they are not held out.
- **LoCoMo.** The BECOS configuration was developed on LoCoMo.
- **BEAM.** Two changes to how BECOS records who said something were tested on conversation 1 before the main runs. Details of the setup are in Appendix B.

**What is not yet separated.** These comparisons measure whole systems. To see what consolidation itself adds, every setup would need the same sessions, the same step and starting code, and the same amount of context. Only the delivered text would differ: search over raw sessions, a flat list of extracted facts, or consolidated facts with status and evidence. The same holds for the mechanisms in sections 4–6, such as the update actions, the settling of proposals and the question types: none is measured on its own here. A future paper will measure the effect of each one separately.

**A source is not proof.** A link lets a reader check a claim, but does not make it true. Recall therefore shows each claim's evidence label beside it, so the agent can tell a requirement or a report from a verified result.

**Privacy.** Sessions can contain sensitive information. A deployment must decide what is captured, how long it is kept, and who can see both the claims and their sources. If a source is removed, someone must decide whether the claims built on it can stay. Correcting a claim is different from asking for personal data to be erased. These are choices for people; a ranking method cannot make them.

**Data and code availability.** The benchmark harness, the mobility tasks with their hidden tests, and the result summaries for the mobility runs are at github.com/becos-ai/becos-benchmark-public. The team questions and per-run results are public at becos.ai/benchmark; the sessions behind them are private.

**Competing interests.** The authors build BECOS and ran every benchmark themselves.

## 12. Conclusion

Code passes easily from one developer to the next. The reasons behind it do not. BECOS keeps those reasons with their status and evidence, and brings them back when a later task needs them. Across five benchmarks, it improved code built over many sessions, answered questions that the other setups did not, avoided repeating long investigations, and kept a developer's rules in view. It adds most where no note, commit or code holds the answer, and it does so with a small fraction of the text. The next step is several people writing code together through one memory, so that each person's agent can build on what the whole team has already learned.

## References

1. Endel Tulving. “Episodic and Semantic Memory.” In *Organization of Memory*, edited by E. Tulving and W. Donaldson. Academic Press, 1972, pp. 381–403.

2. Patrick Lewis et al. Retrieval-Augmented Generation for Knowledge-Intensive NLP Tasks. NeurIPS, 2020.
3. Joon Sung Park et al. Generative Agents: Interactive Simulacra of Human Behavior. UIST, 2023.
4. Charles Packer et al. MemGPT: Towards LLMs as Operating Systems. arXiv, 2023.
5. Prateek Chhikara et al. Mem0: Building Production-Ready AI Agents with Scalable Long-Term Memory. arXiv, 2025.
6. Preston Rasmussen et al. Zep: A Temporal Knowledge Graph Architecture for Agent Memory. arXiv, 2025.
7. Adyasha Maharana et al. Evaluating Very Long-Term Conversational Memory of LLM Agents. ACL, 2024.
8. Haoran Sun, Shaoning Zeng, and Bob Zhang. H-MEM: Hierarchical Memory for High-Efficiency Long-Term Reasoning in LLM Agents. EACL, 2026, pp. 341–350.
9. Kai Li et al. TiMem: Temporal-Hierarchical Memory Consolidation for Long-Horizon Conversational Agents. Findings of ACL, 2026.
10. Han Chen et al. MemForest: Write-Efficient Agent Memory with Decoupled Extraction and Hierarchical View Maintenance. arXiv, 2026.
11. Alireza Rezazadeh et al. Collaborative Memory: Multi-User Memory Sharing in LLM Agents with Dynamic Access Control. arXiv, 2025.
12. Sanjana Pedada, Aditya Dhavala, and Neelraj J. Patil. Shared Selective Persistent Memory for Agentic LLM Systems. REALM Workshop, 2026.
13. Jingyu Liu et al. RAG in a Trenchcoat: When Minimal Memory Is Enough for Agentic Systems, and When It Isn't. NeurIPS, 2026.
14. Nikhil Kumar Jha and Amitoj Cheema. What Moves a LoCoMo Score? Memory Context, Answer Prompts and Judges. arXiv, 2026.
15. Ripon Chandra Malo and Tong Qiu. PROJECTMEM: A Local-First, Event-Sourced Memory and Judgment Layer for AI Coding Agents. arXiv, 2026.
16. Yuvraj Sehgal, Mahsa Panahandeh, and Naser Ezzati-Jivan. Stratum: Repository Memory for Coding Agents Beyond the Context Window. arXiv, 2026.
17. Peiwen Li et al. CoReMem: Cost-Internalizing Memory for Successive Coding Tasks with Prospective Valuation. arXiv, 2026.
18. Liyang Fan et al. VibeMemBench: Evaluating Memory Systems for Coding Agents on Real Repository Coding Tasks. arXiv, 2026.
19. Mohammad Tavakoli, Alireza Salemi, Carrie Ye, Mohamed Abdalla, Hamed Zamani, and J. Ross Mitchell. Beyond a Million Tokens: Benchmarking and Enhancing Long-Term Memory in LLMs. arXiv, 2025.

## Appendix A. Related systems

**Retrieval and agent memory.** Retrieval-augmented generation gives a model outside evidence at answer time [2]. Generative Agents stores experiences and reflects on them [3]. MemGPT moves information between a limited context and outside storage [4]. BECOS builds on these ideas. It adds a record of whether each claim was proposed, decided or verified.

**Systems that update facts.** Mem0 extracts facts from conversations and updates them [5]. Zep tracks when each fact was valid and links it to its source [6].

**Systems that organize memory in layers.** H-MEM arranges memories from general to specific and searches layer by layer [8]. TiMem builds a tree over time, from observations up to a picture of a person [9]. MemForest keeps facts separate from the summaries built on them, and merges facts that say the same thing [10]. BECOS adds each decision's status and evidence.

**Systems that share memory.** Collaborative Memory records where shared fragments came from and uses that to control access [11]. Shared selective persistent memory keeps requirements from code-generating sessions, drops the reasoning, and passes the result to other users [12].

**Memory for coding agents.** Recent systems keep what an agent learns about a project. projectmem logs decisions, attempts and fixes, and a new decision can name the one it replaces [15]. Stratum keeps a repository's structure, its history and what agents found out [16]. CoReMem carries code locations and test commands from one task to the next [17]. VibeMemBench tests whether memory of an agent's earlier fixes helps it solve new issues [18].

BECOS shares these aims. It takes claims from the sessions themselves, records who said each one and its status, and keeps a change apart from a correction (§5). One memory can hold the work of several developers (§9.2).

The table below compares focus and does not say that another system lacks a feature. Section 9.3 compares BECOS with LangMem, Graphiti (Zep's graph engine), Mem0 and Bridge on one public benchmark. Section 9.4 uses LoCoMo [7], which tests memory of personal conversations.

| Work | Main idea, as its authors describe it | What BECOS focuses on |
|---|---|---|
| H-MEM [8] | Layers from general to specific | A claim keeps its status and evidence wherever it is read |
| TiMem [9] | A tree over time, ending in a picture of a person | Software and the decisions about it |
| MemForest [10] | Lasting facts, with summaries rebuilt from a time-ordered tree | Each decision's status, and why a claim should be believed |
| Zep [6] | When a fact was valid, with links to its source | Status as well as time: a recent proposal may never have been adopted |
| Collaborative Memory [11] | Source records used to control access | Source records shown to the reader |
| projectmem [15] | A log of decisions, attempts and fixes, checked before an edit | Claims taken from sessions, with who said them and their status |
| Stratum [16] | A repository's structure, history and findings, served within a visible budget | The decisions behind the code, and the people who made them |
| CoReMem [17] | Locations and commands that save the next task effort | Reasons and rules that a later task must respect |

## Appendix B. BEAM run details

**Data and code.** BEAM at commit `b2da22e`, 100K-token split, conversations 1–20. Where BEAM provides a shorter chat file (conversations 5, 7, 10–12 and 16–20), its code loads that file, and so did we for conversations 10–20. Conversations 1–9 were run earlier on the full chat files; for 5 and 7 these differ by six messages.

**Models.** gpt-4o-mini answered at temperature 0. gpt-4.1-mini judged, as in BEAM's code. Each question was answered four times. BECOS's four answers reuse one recall per question, so its score ranges are narrower than the other setups'.

**Prompts.** BEAM's code asks the whole conversation for the answer only, without explanation. Memory and search are told to answer only from the given context and not to use the model's own knowledge. On two instruction-following questions in conversation 12, this strict prompt made both BECOS and BEAM's search decline to answer in all runs.

**Judge.** BEAM's judge prompt contains the rubric item and the answer, but not the question (BEAM issue #15). This applies to every setup.

**Event ordering.** BEAM's report scores event ordering with a ranking score (Kendall τ). In our runs the step that matches answer lines to rubric events rarely found a match, so the ranking score stayed near its floor for every setup (0.13–0.17). We report the judge's score.

**Answer keys.** Reading the source messages, we found 17 questions in conversations 10–20 whose key does not match the conversation. Most keep a value the user later changed; every memory setup gave the newer value and scored 0. Without these questions, the scores are 0.129 (no memory), 0.386 (whole conversation), 0.428 (BECOS, BEAM's setup) and 0.495 (BECOS, same prompt).

| Conversation | Question | What the key says | What the conversation says |
|---|---|---|---|
| 10 | Knowledge update 0 | 1,350 words (message 64) | 1,800 by message 242 |
| 10 | Temporal 1 | 38 days | 53 days to completing the challenge |
| 11 | Knowledge update 0 | March 27 | Date not in the conversation |
| 12 | Knowledge update 0 | April 22 | Date not in the conversation |
| 12 | Instruction following 1 | A rule | Rubric not taken from the user's rule |
| 13 | Information extraction 0 | 7 series, 4,200 pages (message 26) | 8 series, 4,800 pages (message 60) |
| 13 | Knowledge update 1 | $50 (message 204) | $35 (message 238) |
| 14 | Knowledge update 0 | $75 (message 102) | $70, then $85 |
| 14 | Knowledge update 1 | 30 cupcakes | 24 (message 118); the key's sources do not mention cupcakes |
| 15 | Information extraction 0 | 3 PM (message 34) | Moved to 4 PM (message 56) |
| 15 | Multi-session 0 | Two sizes | Three sizes (message 200) |
| 16 | Preference 0 | Spreadsheet, no apps | User switched to YNAB (messages 166–202) |
| 17 | Abstention 0 | No technique named | Named in message 106 |
| 17 | Temporal 1 | 46 days | April 20 to July 5 is 76 days |
| 18 | Abstention 0 | No information on the advice | Given in messages 254–256 |
| 18 | Abstention 1 | No information on the format | Zoom (message 68), biweekly (message 214) |
| 18 | Knowledge update 0 | 4 hours (message 114) | 5 hours (message 166) |

**Scores per conversation.** Mean of four answers. Conversation 1's no-memory and whole-conversation scores are single runs.

| Conversation | No memory | Whole conversation | BECOS, BEAM's setup | BECOS, same prompt |
|---|---|---|---|---|
| 1 | 0.209 | 0.521 | 0.525 | – |
| 2 | 0.322 | 0.495 | 0.597 | – |
| 3 | 0.158 | 0.422 | 0.364 | – |
| 4 | 0.323 | 0.324 | 0.489 | – |
| 5 | 0.177 | 0.433 | 0.285 | – |
| 6 | 0.215 | 0.337 | 0.404 | – |
| 7 | 0.159 | 0.296 | 0.460 | – |
| 8 | 0.220 | 0.401 | 0.549 | – |
| 9 | 0.189 | 0.424 | 0.494 | – |
| 10 | 0.200 | 0.348 | 0.530 | 0.647 |
| 11 | 0.200 | 0.360 | 0.375 | 0.509 |
| 12 | 0.236 | 0.393 | 0.404 | 0.493 |
| 13 | 0.088 | 0.295 | 0.249 | 0.297 |
| 14 | 0.113 | 0.237 | 0.319 | 0.356 |
| 15 | 0.106 | 0.315 | 0.537 | 0.570 |
| 16 | 0.135 | 0.491 | 0.420 | 0.491 |
| 17 | 0.088 | 0.279 | 0.372 | 0.535 |
| 18 | 0.209 | 0.597 | 0.519 | 0.587 |
| 19 | 0.066 | 0.426 | 0.384 | 0.435 |
| 20 | 0.020 | 0.394 | 0.360 | 0.413 |

### Per ability, conversations 10–20.

| Ability | No memory | Whole conversation | BECOS, BEAM's setup | BECOS, same prompt |
|---|---|---|---|---|
| Abstention | 0.48 | 0.59 | 0.48 | 0.65 |
| Contradiction resolution | 0.00 | 0.12 | 0.12 | 0.22 |
| Event ordering | 0.19 | 0.45 | 0.55 | 0.54 |
| Information extraction | 0.04 | 0.51 | 0.58 | 0.67 |
| Instruction following | 0.09 | 0.39 | 0.37 | 0.48 |
| Knowledge update | 0.05 | 0.30 | 0.32 | 0.44 |
| Multi-session reasoning | 0.10 | 0.56 | 0.56 | 0.59 |
| Preference following | 0.35 | 0.53 | 0.61 | 0.79 |
| Summarization | 0.04 | 0.23 | 0.23 | 0.30 |
| Temporal reasoning | 0.00 | 0.09 | 0.25 | 0.18 |